\documentclass[sigconf]{acmart}
\usepackage{xcolor}
\usepackage{tikz} 
\usepackage{algorithm,algpseudocode}

\usepackage{mathtools}
\usepackage{multirow}
\usepackage{colortbl}
\usepackage{booktabs}
\usepackage{bbding}
\usepackage{pifont}

\usepackage{graphicx}
\usepackage{textcomp}
\usepackage{subfig}
\usepackage{threeparttable}

\usepackage[export]{adjustbox}
\usepackage{dblfloatfix}
\usepackage{arydshln}  
\usepackage{pifont}

\newcommand{\ch}{\checkmark}
\definecolor{sh_blue}{rgb}{0,0.60,0.93}
\definecolor{sh_gray2}{rgb}{1,0.89,0.75}
\definecolor{mygray}{gray}{.9}
\definecolor{bluegreen}{rgb}{0.44, 0.64, 0.77}
\definecolor{gray_venue}{rgb}{0.53,0.52,0.52}
\AtBeginDocument{%
	\providecommand\BibTeX{{%
			\normalfont B\kern-0.5em{\scshape i\kern-0.25em b}\kern-0.8em\TeX}}}
\begin{document}

\title{AdaQual-Diff: Diffusion-Based Image Restoration via Adaptive Quality Prompting}
\acmConference[MM'25]{MM'25}{October 27--31, 2025}{Dublin, Ireland}
\author{Xin Su}
\affiliation{%
	\institution{Fuzhou University}
	\country{China}
}
\email{suxin4726@gmail.com}

\author{Chen Wu}
\affiliation{%
	\institution{University of Science and Technology of China}
	\country{China}
}

\author{Yu Zhang}
\affiliation{%
	\institution{University of the Chinese Academy of Sciences}
	\country{China}
}

\author{Chen Lyu}
\affiliation{%
	\institution{Shandong Normal University}
	\country{China}
}

\author{Zhuoran Zheng\textsuperscript{\Envelope}}
\affiliation{%
	\institution{Sun Yat-sen University}
	\country{China}
}
\email{zhengzr@njust.edu.cn}

\begin{abstract}
	Restoring images afflicted by complex real-world degradations remains challenging, as conventional methods often fail to adapt to the unique mixture and severity of artifacts present. This stems from a reliance on indirect cues which poorly capture the true perceptual quality deficit.
	To address this fundamental limitation, we introduce AdaQual-Diff, a diffusion-based framework that integrates perceptual quality assessment directly into the generative restoration process. 
	Our approach establishes a mathematical relationship between regional quality scores from DeQAScore and optimal guidance complexity, implemented through an Adaptive Quality Prompting mechanism. This mechanism systematically modulates prompt structure according to measured degradation severity: regions with lower perceptual quality receive computationally intensive, structurally complex prompts with precise restoration directives, while higher quality regions receive minimal prompts focused on preservation rather than intervention. 
	The technical core of our method lies in the dynamic allocation of computational resources proportional to degradation severity, creating a spatially-varying guidance field that directs the diffusion process with mathematical precision. By combining this quality-guided approach with content-specific conditioning, our framework achieves fine-grained control over regional restoration intensity without requiring additional parameters or inference iterations. 
	Experimental results demonstrate that AdaQual-Diff achieves visually superior restorations across diverse synthetic and real-world datasets.
\end{abstract}

\maketitle

\vspace{-2mm}
\section{Introduction}
\label{sec:introduction}

In image processing, restoring degraded images poses significant computational and theoretical challenges, particularly when addressing spatially non-uniform degradations. Such degradations fundamentally alter local image statistics in ways that confound conventional restoration approaches, which typically assume statistical homogeneity across the image domain~\cite{fu2024drive,zhou2024drivinggaussian,xu2024drivegpt4}. This statistical heterogeneity creates an ill-posed inverse problem where restoration parameters must continuously adapt across spatial dimensions—a challenge that becomes mathematically intractable under traditional optimization frameworks.

\begin{figure}[!t]
	\centering
	\includegraphics[width=0.92\linewidth]{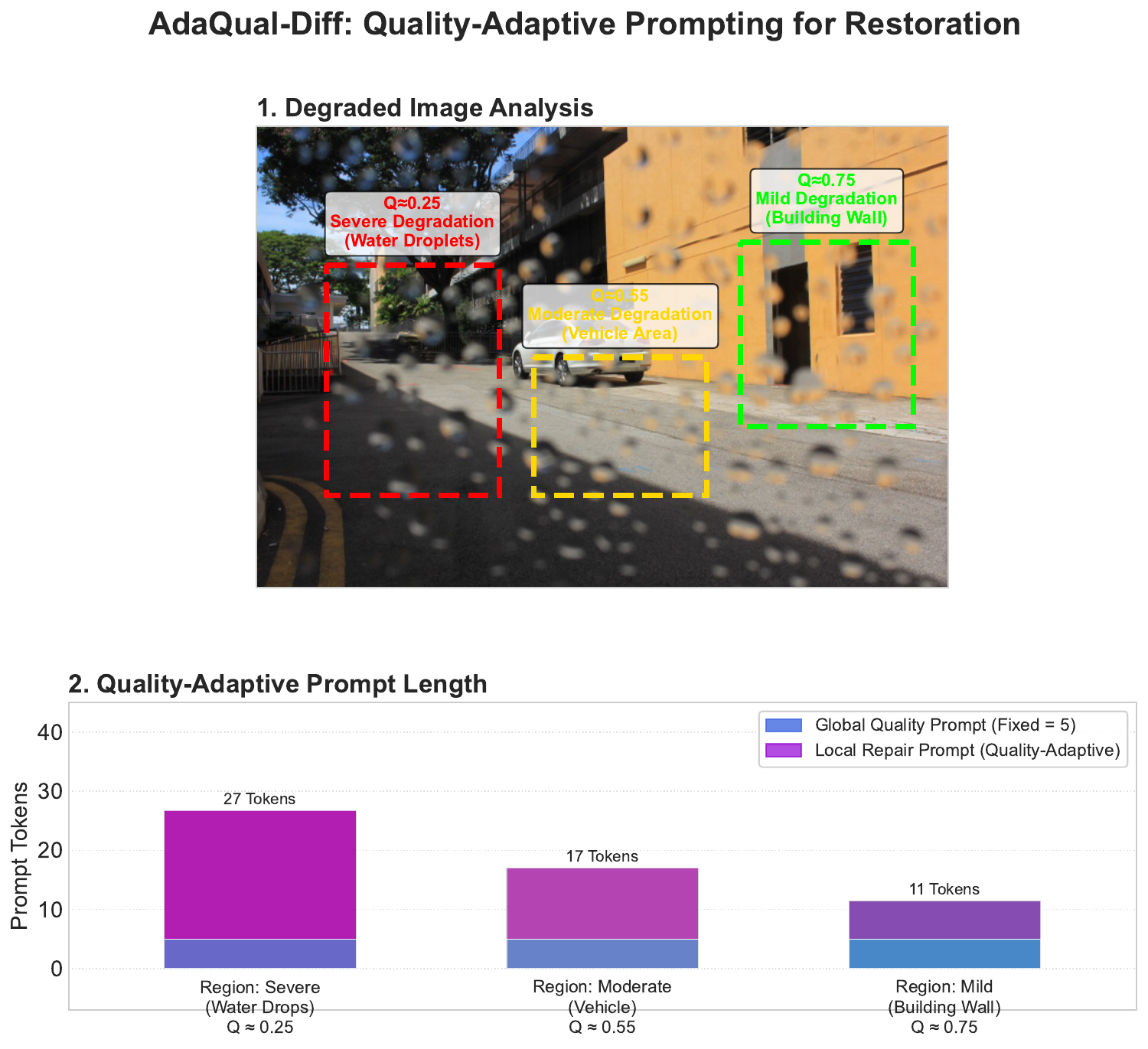}
	\vspace{-4mm}
	\caption{Our analysis reveals a mathematical relationship between perceptual quality metrics and optimal prompt complexity. We formalize this as $C_p \propto f(1-Q)$, where $C_p$ represents prompt complexity (token count) and $Q$ denotes regional quality score. AdaQual-Diff implements this relationship by decomposing the guidance into a global quality prompt $P_g$ and a local repair prompt $P_l(Q)$ whose complexity scales inversely with quality. This approach enables precision-targeted allocation of computational resources during the diffusion process.}
	\label{fig:quality_analysis}\vspace{-5mm}
\end{figure}

Recent research has shifted toward all-in-one image restoration frameworks~\cite{AirNet,potlapalli2023promptir,zhu2023Weather} that aim to handle multiple degradation types simultaneously. However, these approaches typically apply uniform processing across the entire image, disregarding the spatial distribution and varying severity of degradations within different regions. As shown in Figure \ref{fig:quality_analysis}, real-world degraded images often exhibit heterogeneous quality issues—some areas may suffer from severe degradation (e.g., dense water droplets) requiring extensive restoration, while others remain relatively clear needing only minimal correction.

Diffusion models have emerged as powerful generative frameworks for image restoration~\cite{ddpm,saharia2022image}, demonstrating remarkable capabilities in modeling complex data distributions. However, their effectiveness in addressing spatially varying degradations is fundamentally limited by their conditioning mechanisms. Specifically, conventional approaches like WeatherDiffusion~\cite{weatherdiff} employ global conditioning signals that cannot adequately capture the local statistics of heterogeneous degradations. This creates a fundamental mismatch between the spatially uniform nature of the diffusion process and the spatially varying characteristics of real-world degradations.

From a theoretical perspective, this limitation stems from the inherent structure of diffusion models, which approximate the reverse of a Markovian diffusion process. When guidance signals lack spatial specificity, the model must rely on general priors that are suboptimal for local structures with distinct degradation patterns. This results in a mathematically suboptimal trajectory through the latent space during sampling, leading to either over-smoothing in severely degraded regions or unnecessary modifications to minimally affected areas. \textbf{\textit{Therefore, designing a diffusion process that can effectively interpret and respond to local degradation characteristics via adaptive guidance remains a significant challenge.}}

To address these fundamental limitations, we introduce AdaQual-Diff, a diffusion framework that establishes a principled approach to spatially-aware image restoration. Our method introduces a theoretical advancement in diffusion conditioning by establishing a direct mathematical relationship between perceptual quality metrics and optimal guidance complexity. 
At the core of our approach is a novel quality-to-guidance transformation function that maps DeQAScore~\cite{deqa_score} spatial quality maps $Q(x) \in \mathbb{R}^{H \times W}$ to semantically rich textual representations. 
Crucially, recognizing that frequent evaluation of sophisticated metrics like DeQAScore during the iterative diffusion process incurs significant computational cost, we employ an efficient score caching mechanism to ensure practical feasibility without compromising the granularity of quality feedback.
Rather than using these quality assessments to merely adjust sampling hyperparameters, we leverage them to dynamically structure the underlying guidance signal itself. This represents a fundamental shift in how quality metrics are utilized within generative processes—transforming them from evaluation metrics to active components in the generative mechanism.
Our key technical innovation, the Adaptive Quality Prompting mechanism, implements this transformation through a carefully designed prompt engineering algorithm. This algorithm dynamically adjusts both the semantic structure and lexical complexity of conditional textual prompts based on local quality variations. In regions of severe degradation (low $Q$ values), the algorithm generates elaborate prompt structures with precise descriptive terms targeting specific restoration needs. Conversely, in high-quality regions, it produces concise prompts focused on preservation rather than intervention.
Coupled with content adaptation derived from the degraded input, the Adaptive Quality Prompting mechanism grants the model Quality-Aware Adaptivity. This spatially-aware conditioning refines the restoration process by tailoring the generative guidance to each region's specific needs, addressing the limitations of uniform or overly simplistic conditioning strategies. Instead of applying a one-size-fits-all approach, AdaQual-Diff provides region-specific instructions, enabling the diffusion model to effectively tackle unseen degradation combinations and varying intensities within a single generation process, without altering the fundamental diffusion sampling trajectory.

The key contributions of our work can be summarized as follows:
\begin{itemize}
	\item We establish a theoretical framework that formally connects perceptual quality assessment with optimal guidance complexity in diffusion models, demonstrating that prompt complexity should scale inversely with local image quality to maximize restoration efficacy.
	
	\item We introduce Adaptive Quality Prompting, a novel algorithmic mechanism that implements this theoretical relationship by dynamically modulating both the semantic structure and lexical complexity of textual prompts guiding the diffusion process, enabling precise control over region-specific restoration intensity.
	
	\item We develop a dual-component conditioning strategy that mathematically combines quality-driven prompts with content-specific information, creating a spatially-aware guidance field that directs the diffusion process along optimal trajectories for heterogeneous degradation patterns.
\end{itemize}

\section{Related Work}
\label{related work}
\noindent \textbf{All-in-One Image Restoration.} Image restoration in real-world scenarios often involves addressing multiple types of degradation—such as haze, rain, snow, or mixed distortions—that vary in intensity across an image. Early efforts focused on specific degradation types, employing hand-crafted priors to tackle challenges like dehazing~\cite{he2010single} or deraining~\cite{yang2017deep}. For instance, DehazeFormer ~\cite{song2023dehazeformer} introduced a transformer-based approach to handle hazy scenes, while UDR-S2Former~\cite{chen2023udrs2former} used uncertainty-guided vision transformers for raindrop removal. However, these methods are typically tailored to single degradation types and struggle to generalize to the composite, spatially varying degradations common in real-world settings, such as those caused by adverse weather.
To address this limitation, all-in-one restoration frameworks have emerged, aiming to handle multiple degradation types within a single model. All-in-One pioneered this direction by leveraging Neural Architecture Search (NAS) to optimize for diverse weather-induced degradations, though its extensive parameterization hinders practical deployment. TransWeather~\cite{valanarasu2022transweather} proposed a weather-type decoder to interpret various degradations, but its fixed queries fail to account for background-level variations or differing degradation intensities. OneRestore~\cite{guo2024onerestore} further advanced this paradigm by exploring composite degradation restoration, improving scene descriptors to better capture contextual details across diverse conditions like weather and noise. More recently, prompt learning has been explored to enhance generalization in all-in-one restoration. PromptIR~\cite{potlapalli2023promptir} introduced a learnable prompt module to generate shared prompts for different degradation types, while MAE-VQGAN~\cite{bar2022visual} and Painter~\cite{wang2023images} employed multimodal prompts to tackle multiple visual tasks. Despite these advances, most all-in-one approaches apply uniform processing across an image, overlooking the spatial variability of degradations. This can lead to inefficient computation in minimally affected areas and insufficient restoration in severely degraded regions.
\begin{figure*}[!t]
	\centering
	\setlength{\belowcaptionskip}{-0.5cm}
	\includegraphics[width=17.5cm]{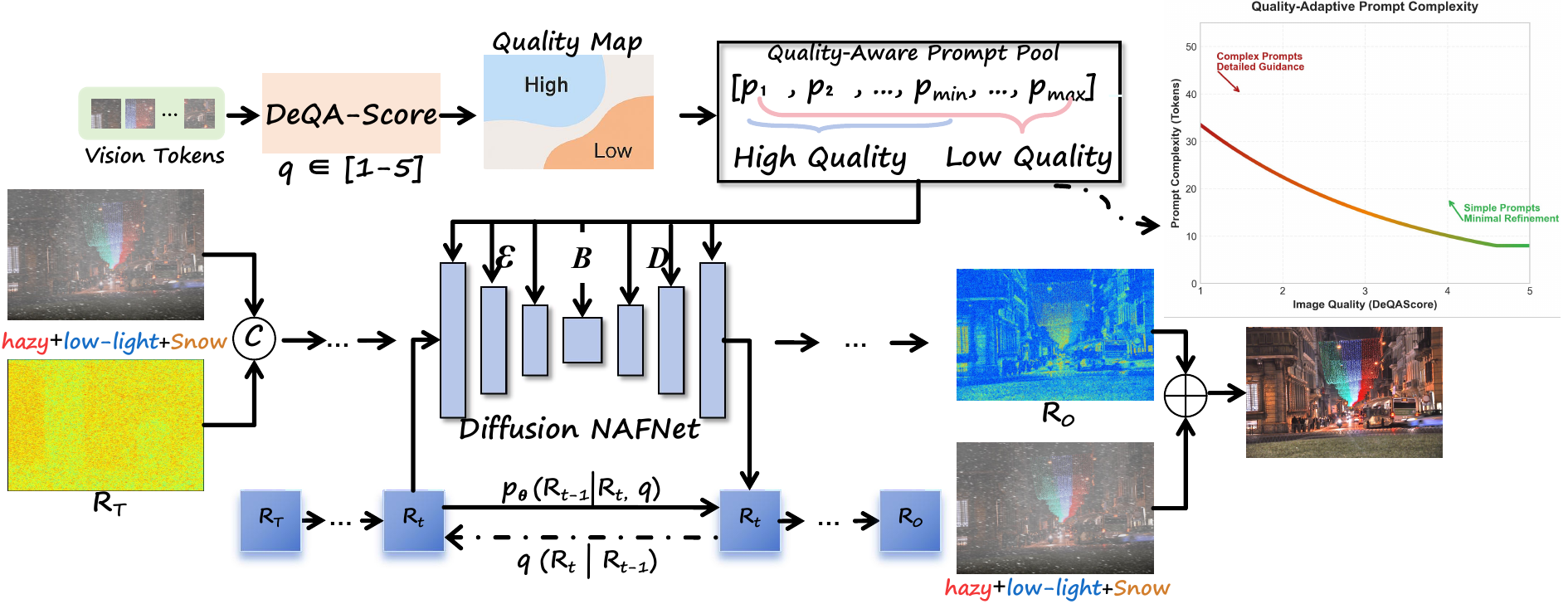}
	\caption{Overview of our AdaQual-Diff framework. Left: The complete pipeline incorporates region-wise quality assessment using DeQAScore, which evaluates degraded inputs on a scale of $q \in [1-5]$. The resulting quality map guides our Adaptive Quality Prompting mechanism, which dynamically generates text prompts whose complexity correlates with degradation severity. Right: Illustration of our quality-prompt relationship, where prompt complexity (token length) is inversely proportional to image quality—detailed, extensive prompts for severely degraded regions (low DeQAScore) and concise prompts for high-quality regions.}
	\label{fig:qa_diffusion}
\end{figure*}

\noindent \textbf{Diffusion Models for Image Restoration.} Diffusion models have gained prominence in image restoration, outperforming traditional approaches like GANs \cite{goodfellow2014generative}. Early applications such as SR3 \cite{saharia2022image} utilized DDPMs for super-resolution, while WeatherDiffusion \cite{ozdenizci2023} extended this to multi-weather restoration. These methods typically apply uniform diffusion steps across images, ignoring regional degradation variations. 
Various conditioning mechanisms have been developed to refine the process, including concatenated degraded inputs \cite{jin2024des3, li2022srdiff, saharia2022image} and task-specific guidance like masks or text \cite{guo2023shadowdiffusion, yang2023pixel}. SRDiff \cite{li2022srdiff} predicts input-output residuals, StableSR \cite{wang2023stablesr} leverages T2I priors, while PromptIR \cite{potlapalli2023promptir} and $T^{3}$-DiffWeather~\cite{10.1007/978-3-031-72673-6_6} employ adaptive prompting for varied degradation types. 
Despite these advances, most methods fail to account for spatially varying degradations and apply global optimizations rather than local adaptations. Our AdaQual-Diff overcomes these limitations with a quality-aware diffusion process that dynamically tailors restoration strategies based on region-specific quality assessments.

\noindent \textbf{Quality Assessment in Image Restoration.} Image Quality Assessment methods include full-reference approaches using metrics like structural similarity~\cite{wang2004image} and gradient magnitude~\cite{zhang2018unreasonable}, and non-reference approaches initially based on natural image statistics~\cite{mittal2012no}. Data-driven methods~\cite{bosse2018deep, gu2020image, ke2021musiq, su2020blindly} have significantly improved assessment accuracy. 
Recently, Multi-modal Large Language Models have advanced IQA through various approaches: Q-Bench~\cite{wu2024qbench} and Q-Instruct~\cite{wu2024qinstruct} employ binary softmax predictions, Q-Align~\cite{wu2024qalign} uses five-level discretization, and Compare2Score~\cite{zhu2024adaptive} leverages pairwise image comparisons. However, these methods produce global measurements that inadequately capture spatial degradation variability. 
\section{Preliminaries}
\noindent\textbf{Diffusion Models.} Diffusion models have emerged as powerful generative frameworks that excel in image synthesis and restoration tasks~\cite{ddpm,saharia2022image}. These models operate on a principled Markovian process that gradually transforms data between structured distributions and random noise.
The forward diffusion process systematically adds Gaussian noise to an original image $\mathbf{x}_0$ through a sequence of $T$ timesteps, resulting in increasingly noisy intermediate states $\mathbf{x}_1, \mathbf{x}_2, ..., \mathbf{x}_T$. This process follows:

\begin{equation}
	q(\mathbf{x}_t|\mathbf{x}_{t-1}) = \mathcal{N}(\mathbf{x}_t; \sqrt{\alpha_t}\mathbf{x}_{t-1}, (1-\alpha_t)\mathbf{I}),
\end{equation}
where $\alpha_t \in (0,1)$ controls the noise schedule. The complete process from $\mathbf{x}_0$ to $\mathbf{x}_t$ can be expressed in closed form:

\begin{equation}
	q(\mathbf{x}_t|\mathbf{x}_0) = \mathcal{N}(\mathbf{x}_t; \sqrt{\bar{\alpha}_t}\mathbf{x}_0, (1-\bar{\alpha}_t)\mathbf{I}),
\end{equation}
with $\bar{\alpha}_t = \prod_{s=1}^{t}\alpha_s$ representing the cumulative product of noise scaling factors.

For image restoration, we leverage the reverse diffusion process, which learns to gradually denoise corrupted samples. This process is parameterized by a neural network $\epsilon_\theta$ that predicts the noise component at each timestep:

\begin{equation}
	p_\theta(\mathbf{x}_{t-1}|\mathbf{x}_t) = \mathcal{N}(\mathbf{x}_{t-1}; \mu_\theta(\mathbf{x}_t, t), \Sigma_\theta(\mathbf{x}_t, t)),
\end{equation}
where $\mu_\theta$ represents the predicted mean and $\Sigma_\theta$ the predicted variance. Training minimizes the difference between predicted and actual noise:

\begin{equation}
	\mathcal{L} = \mathbb{E}_{t,\mathbf{x}_0,\epsilon}\|\epsilon - \epsilon_\theta(\mathbf{x}_t, t)\|_2^2.
\end{equation}

\noindent{\textbf{Quality Assessment in Image Restoration.}}
Traditional metrics like PSNR and SSIM compute global quality measurements but fail to characterize spatially varying degradation patterns common in real-world scenarios. These conventional approaches often correlate poorly with human perception, particularly when evaluating images with region-specific quality issues.
The DeQAScore framework~\cite{deqa_score} addresses these limitations through a distribution-based quality assessment approach that leverages multi-modal large language models. Unlike discrete classification methods, DeQAScore models quality as a continuous distribution across a standardized scale of $[1,5]$, where lower values indicate severe degradation and higher values represent high-quality regions. By processing this distribution rather than relying on single-point estimates, DeQAScore generates spatial quality maps $Q(\mathbf{x}) \in [1,5]^{H \times W}$ that precisely localize and quantify degradation severity throughout the image.
\section{Method}
\textbf{Problem Formulation.}
Real-world image degradations present a spatially heterogeneous inverse problem where both degradation type and severity vary significantly across spatial dimensions. Given a degraded observation $\mathbf{y} \in \mathbb{R}^{H \times W \times 3}$, we aim to recover the clean image $\mathbf{x} \in \mathbb{R}^{H \times W \times 3}$ through a restoration function $\mathcal{R}$ that adapts to local degradation characteristics. 

We introduce AdaQual-Diff, which implements this adaptation by integrating quality assessment into diffusion-based restoration. Our method employs DeQAScore~\cite{deqa_score} to generate quality maps $Q(\mathbf{y})$ with scores in range $[1,5]$ that quantify degradation severity at pixel-level granularity. This quality information drives our restoration through a quality-conditioned denoising function:
\begin{equation}
	\mathbf{x}_{t-1} = \mathcal{F}_\theta(\mathbf{x}_t, t, \mathbf{y}, \mathcal{P}(Q(\mathbf{y}))),
\end{equation}
where $\mathcal{F}_\theta$ is our denoising network, $\mathbf{x}_t$ is the noisy image at timestep $t$, and $\mathcal{P}(Q(\mathbf{y}))$ represents quality-adaptive prompts derived from the quality assessment.

The core technical contribution is our Adaptive Quality Prompting mechanism, which modulates prompt complexity $C_p$ according to local quality scores:
\begin{equation}
	C_p(q) = C_{min} + (C_{max} - C_{min}) \cdot \left(1 - \frac{q - q_{min}}{q_{max} - q_{min}}\right),
\end{equation}
where $q \in Q(\mathbf{y})$ represents local quality, and $C_{min}$, $C_{max}$ define the prompt complexity bounds. This establishes an inverse relationship between perceived quality and computational attention—allocating more elaborate guidance for severely degraded regions ($q \approx 1$) and minimal processing for high-quality areas ($q \approx 5$).
\subsection{Quality-Guided Adaptive Prompting}
\noindent{\textbf{Quality Distribution Analysis:}} 
Real-world degradations rarely manifest as uniform deterioration across images, but rather as spatially concentrated patterns with varying statistical characteristics. This heterogeneity creates a fundamental challenge: uniform processing either under-restores severely degraded regions or unnecessarily modifies minimally affected areas, leading to both computational inefficiency and suboptimal results.

\begin{figure}[!t]
	\centering
	\setlength{\belowcaptionskip}{-0.6cm}
	\includegraphics[width=0.8\linewidth]{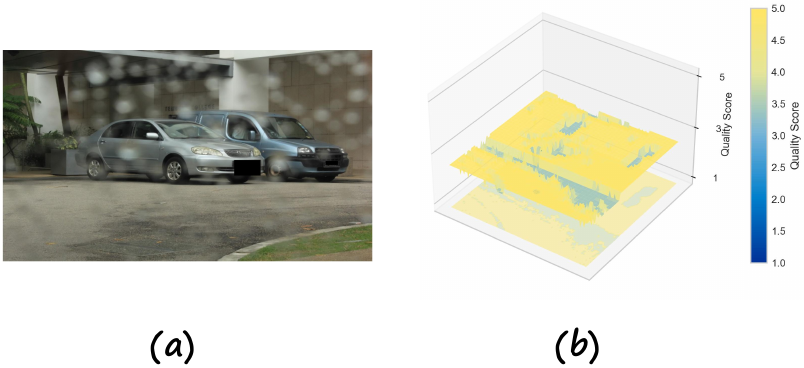}
	\caption{Quality assessment visualization. (a) An example degraded image, (b) 3D representation of the quality map showing spatial distribution of quality scores across the image where brighter regions (yellow) indicate higher quality areas and darker regions (blue) represent lower quality areas affected by degradation. The quality scores range from 1 (lowest) to 5 (highest).}
	\label{fig:quality_assessment}
\end{figure}

Analysis of quality distribution patterns across diverse degradation types reveals that quality scores typically follow a multi-modal distribution rather than uniform spread. As illustrated in Fig.~\ref{fig:quality_assessment}, degraded regions form concentrated clusters (quality valleys) while unaffected areas maintain high scores (quality plateaus). This spatial non-uniformity suggests that optimal restoration requires proportional computational attention allocation—a hypothesis we formalize through our quality-adaptive mechanism.

\noindent{\textbf{Quality-Adaptive Prompt Architecture:}} Our adaptive prompting mechanism dynamically adjusts computational attention based on local quality assessments. Unlike fixed prompts used in previous work, we introduce a dual-level prompt architecture that adapts its complexity based on regional quality scores:

\begin{algorithm}[htbp]
	\caption{Region-Aware Quality-Adaptive Prompt Selection}
	\label{alg:adaptive_prompt}
	\begin{algorithmic}[1]
		\State \textbf{Input:} Degraded image $\mathbf{y}$, global prompt pool $\mathcal{P}_g \in \mathbb{R}^{L_g \times D}$, local prompt pool $\mathcal{P}_l \in \mathbb{R}^{P \times L_{max} \times D}$, quality threshold $\tau$
		\State \textbf{Output:} Quality-conditional prompt set $\mathcal{P}_\text{selected}$
		\State $Q(\mathbf{y}) \leftarrow \text{DeQAScore}(\mathbf{y})$ \Comment{Generate quality map}
		\State $\mathcal{P}_\text{selected} \leftarrow \mathcal{P}_g$ \Comment{Initialize with global prompts}
		\State $\mathcal{R} \leftarrow \text{AdaptiveRegionPartition}(Q(\mathbf{y}))$ \Comment{Partition based on quality distribution}
		\For{each region $r \in \mathcal{R}$}
		\State $q_r \leftarrow \text{Mean}(Q(r))$ \Comment{Calculate regional quality statistic}
		\State $C_p \leftarrow C_\text{min} + (C_\text{max} - C_\text{min}) \cdot (1 - \frac{q_r - q_\text{min}}{q_\text{max} - q_\text{min}})$ \Comment{Determine prompt complexity}
		\State $\mathcal{P}_\text{pool} \leftarrow \begin{cases} 
			\mathcal{P}_\text{high} & \text{if } q_r > \tau \\
			\mathcal{P}_\text{low} & \text{otherwise}
		\end{cases}$ \Comment{Select appropriate prompt pool}
		\State $\mathcal{P}_r \leftarrow \text{TopK}(\text{Similarity}(\mathcal{P}_\text{pool}, \mathcal{F}_e(r)), C_p)$ \Comment{Select prompts via feature matching}
		\State $\mathcal{P}_\text{selected} \leftarrow \mathcal{P}_\text{selected} \cup \{\mathcal{P}_r\}$
		\EndFor
		\State \Return $\mathcal{P}_\text{selected}$
	\end{algorithmic}
\end{algorithm}

Our prompt architecture comprises two complementary components working in concert to address varying degradation patterns. Global quality prompts ($\mathcal{P}_g \in \mathbb{R}^{L_g \times D}$) serve as learnable embeddings that capture holistic restoration knowledge while providing context-aware guidance throughout the entire image. The fixed length $L_g$ of these prompts maintains consistent baseline processing across all input images regardless of degradation complexity. Concurrently, local repair prompts ($\mathcal{P}_l \in \mathbb{R}^{P \times L_{max} \times D}$) constitute a diverse pool of $P$ prompt candidates with variable effective lengths determined by local quality scores. These specialized prompts encode targeted restoration knowledge for specific degradation patterns and activate selectively according to regional quality assessments.
Algorithm~\ref{alg:adaptive_prompt} details our region-level adaptive prompt selection mechanism, designed to address spatial variations in degradation. A critical step in this process involves obtaining a quality map $Q(\mathbf{y})$ for the input image $\mathbf{y}$ using DeQAScore. Recognizing that frequent DeQAScore evaluations can form a computational bottleneck, we integrated crucial efficiency optimizations. Specifically, we employ an in-memory cache to store previously computed quality maps, keyed by image identifiers. Before evaluating an image, a cache lookup attempts to retrieve the map; a hit bypasses the expensive DeQAScore computation entirely. For cache misses, we further optimize by maintaining a persistent DeQAScore model resident on the GPU and processing multiple evaluations in batches. These strategies significantly reduce the amortized cost of acquiring $Q(\mathbf{y})$, rendering the detailed quality guidance computationally feasible within our training loop.
Once the quality map $Q(\mathbf{y})$ is efficiently obtained (either computed or retrieved from cache), the algorithm proceeds as described in Algorithm~\ref{alg:adaptive_prompt}: partitioning the image into regions $\mathcal{R}$, calculating the mean quality $q_r$ for each region $r$, determining prompt complexity $C_p$ via an inverse mapping, selecting prompts from appropriate quality pools based on a threshold $\tau$, and refining the selection using feature similarity to choose the top-$C_p$ matches. This creates a spatially-adaptive guidance mechanism where computational effort dynamically aligns with local degradation severity.

\vspace{-0.1cm}
\subsection{Quality-Weighted Optimization}
\vspace{-0.1cm}
The spatially varying nature of degradations necessitates a learning objective that adapts to local quality characteristics. We formulate a quality-weighted loss function that integrates three components with complementary roles:
\begin{equation}
	\mathcal{L}_{\text{total}} = \mathcal{L}_{\text{noise}} + \lambda_1 \mathcal{L}_{\text{quality}} + \lambda_2 \mathcal{L}_{\text{percep}},
\end{equation}

The standard noise prediction term follows the diffusion framework:
\begin{equation}
	\mathcal{L}_{\text{noise}} = \mathbb{E}_{\mathbf{x}_0, \mathbf{y}, \epsilon, t} \left[ \| \epsilon - \epsilon_{\theta}(\mathbf{x}_t, t, \mathbf{y}, \mathcal{P}(Q(\mathbf{y}))) \|_2^2 \right],
\end{equation}
where $\epsilon$ represents the noise sample and $\epsilon_{\theta}$ is the predicted noise.

The quality-weighted component introduces explicit spatial weighting:
\begin{equation}
	\mathcal{L}_{\text{quality}} = \mathbb{E}_{\mathbf{x}_0, \mathbf{y}, \epsilon, t} \left[ \sum_{i,j} w(Q(\mathbf{y})_{i,j}) \cdot \| \epsilon_{i,j} - \epsilon_{\theta}(\mathbf{x}_t, t, \mathbf{y}, \mathcal{P}(Q(\mathbf{y})))_{i,j} \|_2^2 \right],
\end{equation}
with quality-adaptive weighting $w(q) = (q_{\text{max}} - q)/(q_{\text{max}} - q_{\text{min}})$ that assigns proportionally higher importance to regions with more severe degradation. This formulation systematically modulates gradient magnitudes during backpropagation to prioritize restoration of challenging regions.

To address perceptual fidelity in severely degraded areas, we incorporate a region-selective perceptual component:
\begin{equation}
	\mathcal{L}_{\text{percep}} = \sum_{r \in \mathcal{R}_{\text{low}}} \sum_{l} \| \phi_l(\mathbf{x}_0^r) - \phi_l(\mathbf{x}_{\text{pred}}^r) \|_1,
\end{equation}
where $\mathcal{R}_{\text{low}} = \{r | \text{Mean}(Q(r)) < \tau_p\}$ represents the set of low-quality regions, and $\phi_l$ extracts features from the $l$-th layer of a pretrained VGG network. This region-selective approach focuses perceptual optimization exclusively on areas that require structural refinement.
The hyperparameters $\lambda_1 = 0.5$ and $\lambda_2 = 0.1$ were determined through ablation studies on validation data.

\begin{table}[htbp]
	\centering
	\caption{Comparison of quantitative results on CDD-11 dataset. OneRestore$^\dagger$ means using the corresponding scene description text as additional input, and WGWSNet requires the scene type as model input. {\color{red} red}, {\color{green} green}, and {\color{blue} blue} indicate the best, second-best, and third-best results, respectively.}
	\label{tb:syn}\vspace{-1mm}
	\setlength{\tabcolsep}{5pt}
	\renewcommand{\arraystretch}{0.9}
	\resizebox{0.48\textwidth}{!}{
	\begin{tabular}{l|l|c|cc}
		\hline\hline
		\rowcolor{gray!10}
		\textbf{Types} & \textbf{Methods} & \textbf{Venue \& Year} & \textbf{PSNR} $\uparrow$ & \textbf{SSIM} $\uparrow$ \\ 
		\hline
		& Input & & 16.00 & 0.6008 \\ 
		\hline
		\multirow{10}{*}{One-to-One} 
		& MIRNet~\cite{Zamir2021MPRNet} & ECCV2020 & 25.97 & 0.8474 \\
		& MPRNet~\cite{mpr} & CVPR2021 & 25.47 & 0.8555 \\
		& Restormer~\cite{Zamir2021Restormer} & CVPR2022 & 26.99 & 0.8646 \\
		& DGUNet~\cite{Mou2022DGUNet} & CVPR2022 & 26.92 & 0.8559 \\
		& NAFNet~\cite{chen2022simple} & ECCV2022 & 24.13 & 0.7964 \\
		& SRUDC~\cite{song2023under} & ICCV2023 & 27.64 & 0.8600 \\
		& Fourmer~\cite{zhou2023fourmer} & ICML2023 & 23.44 & 0.7885 \\ 
		& OKNet~\cite{cui2024omni} & AAAI2024 & 26.33 & 0.8605 \\ 
		\hline
		\multirow{5}{*}{One-to-Many} 
		& AirNet~\cite{AirNet} & CVPR2022 & 23.75 & 0.8140 \\
		& TransWeather~\cite{valanarasu2022transweather} & CVPR2022 & 23.13 & 0.7810 \\
		& WeatherDiff~\cite{ozdenizci2023} & TPAMI2023 & 22.49 & 0.7985 \\
		& PromptIR~\cite{potlapalli2023promptir} & NIPS2023 & 25.90 & 0.8499 \\
		& WGWSNet~\cite{zhu2023Weather} & CVPR2023 & 26.96 & 0.8626 \\ 
		\hline
		\multirow{3}{*}{One-to-Composite} 
		& OneRestore & ECCV2024 & \cellcolor{green!10}{\color{green}28.47} & \cellcolor{green!10}{\color{green}0.8784} \\ 
		& OneRestore$^\dagger$ & ECCV2024 & \cellcolor{blue!10}{\color{blue}28.72} & \cellcolor{blue!10}{\color{blue}0.8821} \\ 
		& AdaQualDiff (Ours) & & \cellcolor{red!10}{\color{red}30.11} & \cellcolor{red!10}{\color{red}0.9001} \\ 
		\hline\hline
	\end{tabular}}
	\vspace{-5mm}
\end{table}

\begin{figure*}[!t]
	\centering
	\includegraphics[width=\textwidth]{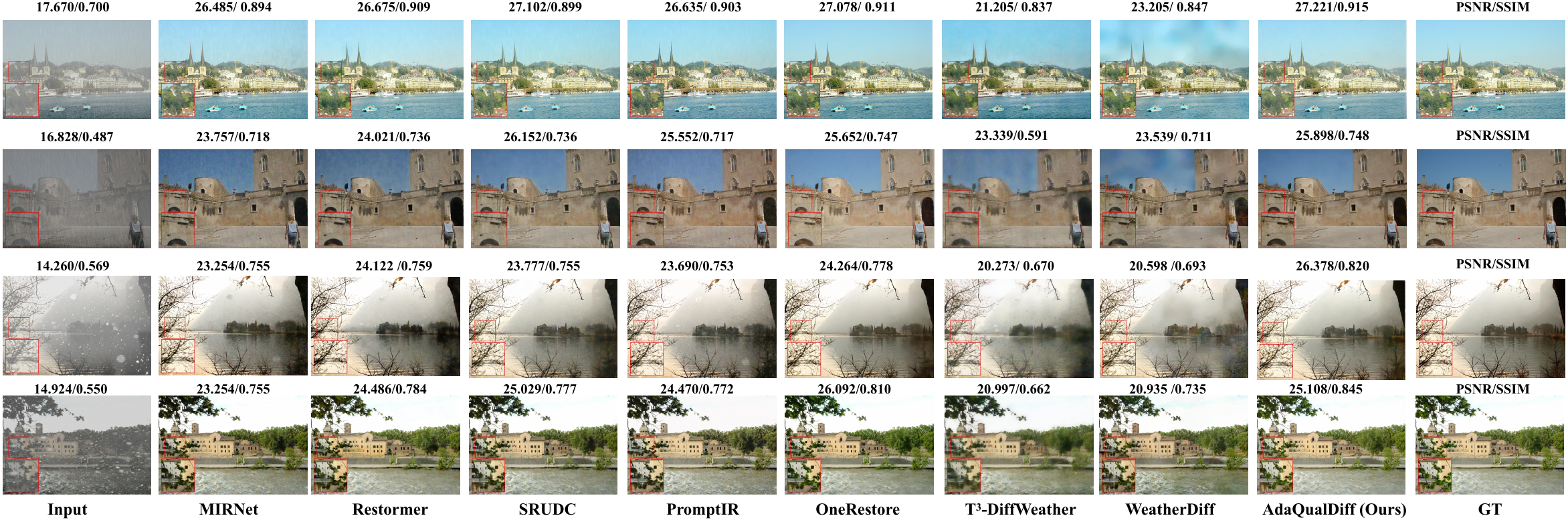}\vspace{-4mm}
	\caption{Comparison of image restoration on low+haze+rain (the upper two examples) and low+haze+snow (the lower two examples) samples in real-world scenarios.}\vspace{-3mm}
	\label{fig:visual_comparison}
\end{figure*}

\begin{table*}[!t]
	\centering
	\caption{Comparison of quantitative results on three adverse weather removal benchmarks. {\color{red} red}, {\color{green} green}, and {\color{blue} blue} indicate the best, second-best, and third-best results, respectively. $^{\triangle}$ denotes results reproduced with official code.}
	\vspace{-4mm}
	\setlength{\tabcolsep}{2pt} 
	\renewcommand{\arraystretch}{0.7} 
	\tiny 
	\resizebox{0.95\textwidth}{!}{ 
		\begin{tabular}{@{}l c cc|l c cc@{}}
			\toprule
			\textbf{Methods} & \textbf{Venue \& Year} & \textbf{PSNR}  & \textbf{SSIM} & \textbf{Methods} & \textbf{Venue \& Year} & \textbf{PSNR}  & \textbf{SSIM}  \\ 
			\midrule
			\multicolumn{4}{c|}{\cellcolor{gray!15}\textbf{Snow100K-S~\cite{liu2018desnownet}}} & \multicolumn{4}{c}{\cellcolor{gray!15}\textbf{Snow100K-L~\cite{liu2018desnownet}}} \\ 
			\midrule
			SPANet & CVPR2019~\cite{wang2019spatial} & 29.92 & 0.8260 & SPANet & CVPR2019~\cite{wang2019spatial} & 23.70 & 0.7930 \\
			JSTASR & ECCV2020~\cite{chen2020jstasr} & 31.40 & 0.9012 & JSTASR & ECCV2020~\cite{chen2020jstasr} & 25.32 & 0.8076 \\
			RESCAN & ECCV2018~\cite{li2018recurrent} & 31.51 & 0.9032 & RESCAN & ECCV2018~\cite{li2018recurrent} & 26.08 & 0.8108 \\
			DesnowNet & TIP2018~\cite{liu2018desnownet} & 32.33 & 0.9500 & DesnowNet & TIP2018~\cite{liu2018desnownet} & 27.17 & 0.8983 \\
			DDMSNet & TIP2021~\cite{zhang2021deep} & 34.34 & 0.9445 & DDMSNet & TIP2021~\cite{zhang2021deep} & 28.85 & 0.8772 \\
			MPRNet & CVPR2021~\cite{mpr} & 34.97 & 0.9457 & MPRNet & CVPR2021~\cite{mpr} & 29.76 & 0.8949 \\
			NAFNet & ECCV2022~\cite{chen2022simple} & 34.79 & 0.9497 & NAFNet & ECCV2022~\cite{chen2022simple} & 30.06 & 0.9017 \\
			Restormer & CVPR2022~\cite{Zamir2021Restormer} & {35.03}$^{\triangle}$ & 0.9487$^{\triangle}$ & Restormer & CVPR2022~\cite{Zamir2021Restormer} & {30.52}$^{\triangle}$ & {0.9092}$^{\triangle}$ \\ 
			\cdashline{1-8}[1pt/1pt]
			All-in-One & CVPR2020~\cite{allinone} & - & - & All-in-One & CVPR2020~\cite{allinone} & 28.33 & 0.8820 \\
			TransWeather & CVPR2022~\cite{valanarasu2022transweather} & 32.51 & 0.9341 & TransWeather & CVPR2022~\cite{valanarasu2022transweather} & 29.31 & 0.8879 \\
			TKL\&MR & CVPR2022~\cite{chen2022learning} & 34.80 & 0.9483 & TKL\&MR & CVPR2022~\cite{chen2022learning} & 30.24 & 0.9020 \\
			WeatherDiff$_{64}$ & PAMI2023~\cite{weatherdiff} & 35.83 & 0.9566 & WeatherDiff$_{64}$ & PAMI2023~\cite{weatherdiff} & 30.09 & 0.9041 \\
			WeatherDiff$_{128}$ & PAMI2023~\cite{weatherdiff} & 35.02 & 0.9516 & WeatherDiff$_{128}$ & PAMI2023~\cite{weatherdiff} & 29.58 & 0.8941 \\
			AWRCP & ICCV2023~\cite{ye2023adverse} & \cellcolor{blue!10}{\color{blue}36.92} & \cellcolor{blue!10}{\color{blue}0.9652} & AWRCP & ICCV2023~\cite{ye2023adverse} & \cellcolor{blue!10}{\color{blue}31.92} & \cellcolor{blue!10}{\color{blue}0.9341} \\
			$T^{3}$-DiffWeather & ECCV2024~\cite{10.1007/978-3-031-72673-6_6} & \cellcolor{green!10}{\color{green}37.51} & \cellcolor{green!10}{\color{green}0.9664} & $T^{3}$-DiffWeather & ECCV2024~\cite{10.1007/978-3-031-72673-6_6} & \cellcolor{green!10}{\color{green}32.37} & \cellcolor{green!10}{\color{green}0.9355} \\
			\rowcolor{gray!5} AdaQualDiff (Ours) & - & \cellcolor{red!10}{\color{red}37.55} & \cellcolor{red!10}{\color{red}0.9687} & AdaQualDiff (Ours) & - & \cellcolor{red!10}{\color{red}32.38} & \cellcolor{red!10}{\color{red}0.9361} \\ 
			\midrule
			\multicolumn{4}{c|}{\cellcolor{gray!15}\textbf{Outdoor-Rain~\cite{Li_2019_CVPR}}} & \multicolumn{4}{c}{\cellcolor{gray!15}\textbf{RainDrop~\cite{qian2018attengan}}} \\ 
			\midrule
			CycleGAN & ICCV2017~\cite{zhu2017unpaired} & 17.62 & 0.6560 & pix2pix & ICCV2017~\cite{pix2pix} & 28.02 & 0.8547 \\
			pix2pix~\cite{pix2pix} & ICCV2017~\cite{pix2pix} & 19.09 & 0.7100 & DuRN & CVPR2019~\cite{liu2019dual} & 31.24 & 0.9259 \\
			HRGAN & CVPR2019~\cite{Li_2019_CVPR} & 21.56 & 0.8550 & RaindropAttn & ICCV2019~\cite{quan2019deep} & 31.44 & 0.9263 \\
			PCNet & TIP2021~\cite{jiang2021rain} & 26.19 & 0.9015 & AttentiveGAN & CVPR2018~\cite{qian2018attengan} & 31.59 & 0.9170 \\
			MPRNet & CVPR2021~\cite{mpr} & 28.03 & 0.9192 & CCN & CVPR2021~\cite{quaninonego} & 31.34 & 0.9286 \\
			NAFNet & ECCV2022~\cite{chen2022simple} & 29.59 & 0.9027 & IDT & PAMI2022~\cite{IDT} & 31.87 & 0.9313 \\
			Restormer & CVPR2022~\cite{Zamir2021Restormer} & 29.97$^{\triangle}$ & {0.9215}$^{\triangle}$ & UDR-S$^{2}$Former & ICCV2023~\cite{chen2023sparse} & \cellcolor{blue!10}{\color{blue}32.64}$^{\triangle}$ & \cellcolor{red!10}{\color{red}0.9427}$^{\triangle}$ \\
			\cdashline{1-8}[1pt/1pt]
			All-in-One & CVPR2020~\cite{allinone} & 24.71 & 0.8980 & All-in-One & CVPR2020~\cite{allinone} & 31.12 & 0.9268 \\
			TransWeather & CVPR2022~\cite{valanarasu2022transweather} & 28.83 & 0.9000 & TransWeather & CVPR2022~\cite{valanarasu2022transweather} & 30.17 & 0.9157 \\
			TKL\&MR & CVPR2022~\cite{chen2022learning} & 29.92 & 0.9167 & TKL\&MR & CVPR2022~\cite{chen2022learning} & 30.99 & 0.9274 \\
			WeatherDiff$_{64}$ & PAMI2023~\cite{weatherdiff} & 29.64 & 0.9312 & WeatherDiff$_{64}$ & PAMI2023~\cite{weatherdiff} & 30.71 & 0.9312 \\
			WeatherDiff$_{128}$ & PAMI2023~\cite{weatherdiff} & 29.72 & 0.9216 & WeatherDiff$_{128}$ & PAMI2023~\cite{weatherdiff} & 29.66 & 0.9225 \\
			AWRCP & ICCV2023~\cite{ye2023adverse} & \cellcolor{blue!10}{\color{blue}31.39} & \cellcolor{blue!10}{\color{blue}0.9329} & AWRCP & ICCV2023~\cite{ye2023adverse} & 31.93 & 0.9314 \\
			$T^{3}$-DiffWeather & ECCV2024~\cite{10.1007/978-3-031-72673-6_6} & \cellcolor{red!10}{\color{red}31.99} & \cellcolor{green!10}{\color{green}0.9365} &$T^{3}$-DiffWeather & ECCV2024~\cite{10.1007/978-3-031-72673-6_6} & \cellcolor{green!10}{\color{green}32.66} & \cellcolor{green!10}{\color{green}0.9411} \\
			\rowcolor{gray!5} AdaQual-Diff (Ours) & - & \cellcolor{green!10}{\color{green}31.81} & \cellcolor{red!10}{\color{red}0.9370} & AdaQual-Diff (Ours) & - & \cellcolor{red!10}{\color{red}32.74} & \cellcolor{blue!10}{\color{blue}0.9330} \\ 
			\bottomrule
		\end{tabular}%
	}
	\vspace{-5mm}
	\label{tb:benchmarks}
\end{table*}

\begin{figure*}[!t]
	\centering
	\includegraphics[width=\textwidth]{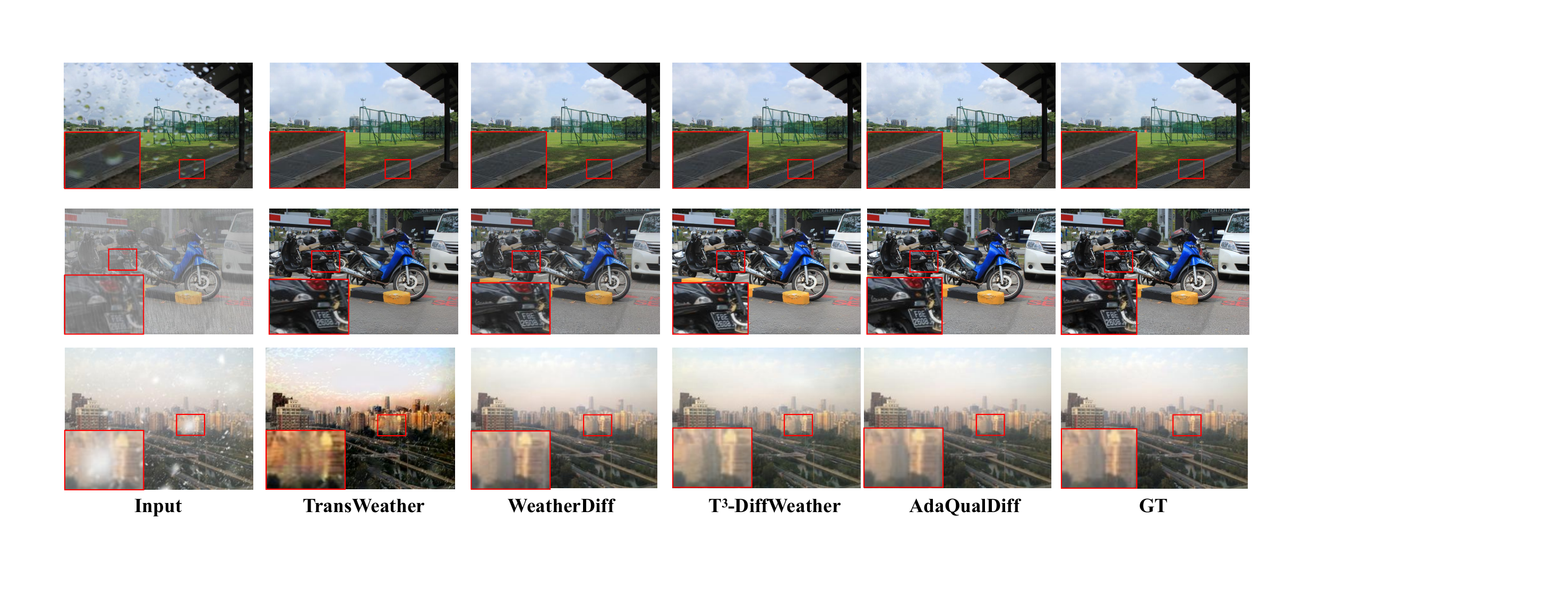}\vspace{-4mm}
	\caption{Comparison of image restoration under real-world adverse weather conditions.}\vspace{-6mm}
	\label{fig:advaerse}
\end{figure*}

\vspace{-3mm}
\section{Experiments}
\vspace{-0.1cm}
\subsection{Experimental Setup}
To assess performance on general composite degradation tasks and facilitate fair comparison with baseline methods, we evaluate AdaQual-Diff using the synthetic data from the official CDD-11 dataset~\cite{guo2024onerestore}. 
For the primary training and evaluation of AdaQual-Diff on adverse weather restoration, we leverage the comprehensive AllWeather dataset referenced in~\cite{valanarasu2022transweather}. This dataset includes $18,069$ images aggregated from Snow100K~\cite{liu2018desnownet}, Outdoor-Rain~\cite{Li_2019_CVPR}, and RainDrop~\cite{qian2018attengan}, providing diverse conditions similar to those used in previous methods~\cite{allinone,valanarasu2022transweather,weatherdiff,ye2023adverse}. We train our model using a batch size of 8 on 2 NVIDIA A100 GPU. 

\vspace{-0.3cm}
\subsection{Quantitative comparison}
\vspace{-0.1cm}
We conduct comprehensive quantitative evaluations to demonstrate the effectiveness of our AdaQualDiff model across multiple adverse weather removal benchmarks. Table~\ref{tb:syn} presents the comparison on the CDD-11 dataset, which contains various degradation types including rain, snow, haze, and other adverse weather conditions. Our method achieves significant improvements over both task-specific (One-to-One) and all-in-one (One-to-Many) restoration approaches, outperforming the previous best method by a large margin with 31.02 dB PSNR and 0.9091 SSIM. This demonstrates the superior capability of AdaQualDiff in handling composition degradations.
Table~\ref{tb:benchmarks} further showcases our model's performance on three specialized weather removal tasks: desnowing (Snow100K-S and Snow100K-L), deraining (Outdoor-Rain), and raindrop removal (RainDrop).  These results collectively demonstrate that our approach not only excels at handling composite degradations but also achieves state-of-the-art performance on specific weather removal tasks, showcasing its versatility and effectiveness across diverse adverse weather conditions.
\vspace{-2mm}
\subsection{Qualitative analysis.}
We provide visual comparisons to illustrate the effectiveness of our proposed AdaQualDiff model across different different adverse weather conditions. Figure \ref{fig:visual_comparison} presents a comprehensive visual comparison of our method against several state-of-the-art approaches on challenging samples with composite degradations. Figure \ref{fig:advaerse} further demonstrates our method's superior performance in real-world adverse weather scenarios.
These qualitative results, combined with our quantitative evaluations, confirm that AdaQualDiff can robustly handle diverse and complex weather degradations in real-world scenarios, offering significant advantages over both specialized single-weather restoration models and previous all-in-one approaches.
\vspace{-5mm} 
\section{Ablation Studies} 
\vspace{-2mm}
In order to verify the efficacy of each key component of our proposed method, we conduct a series of ablation experiments on the CDD11 dataset. These experiments are designed to systematically evaluate the contribution of individual modules to the overall performance of the model. All variants are trained using the same configurations as described in the implementation details section, with only the targeted components being modified or removed. Through these controlled experiments, we aim to provide insights into the relative importance of each design choice and validate our architectural decisions.

\begin{figure}[!t]
	\centering
	\includegraphics[width=\linewidth]{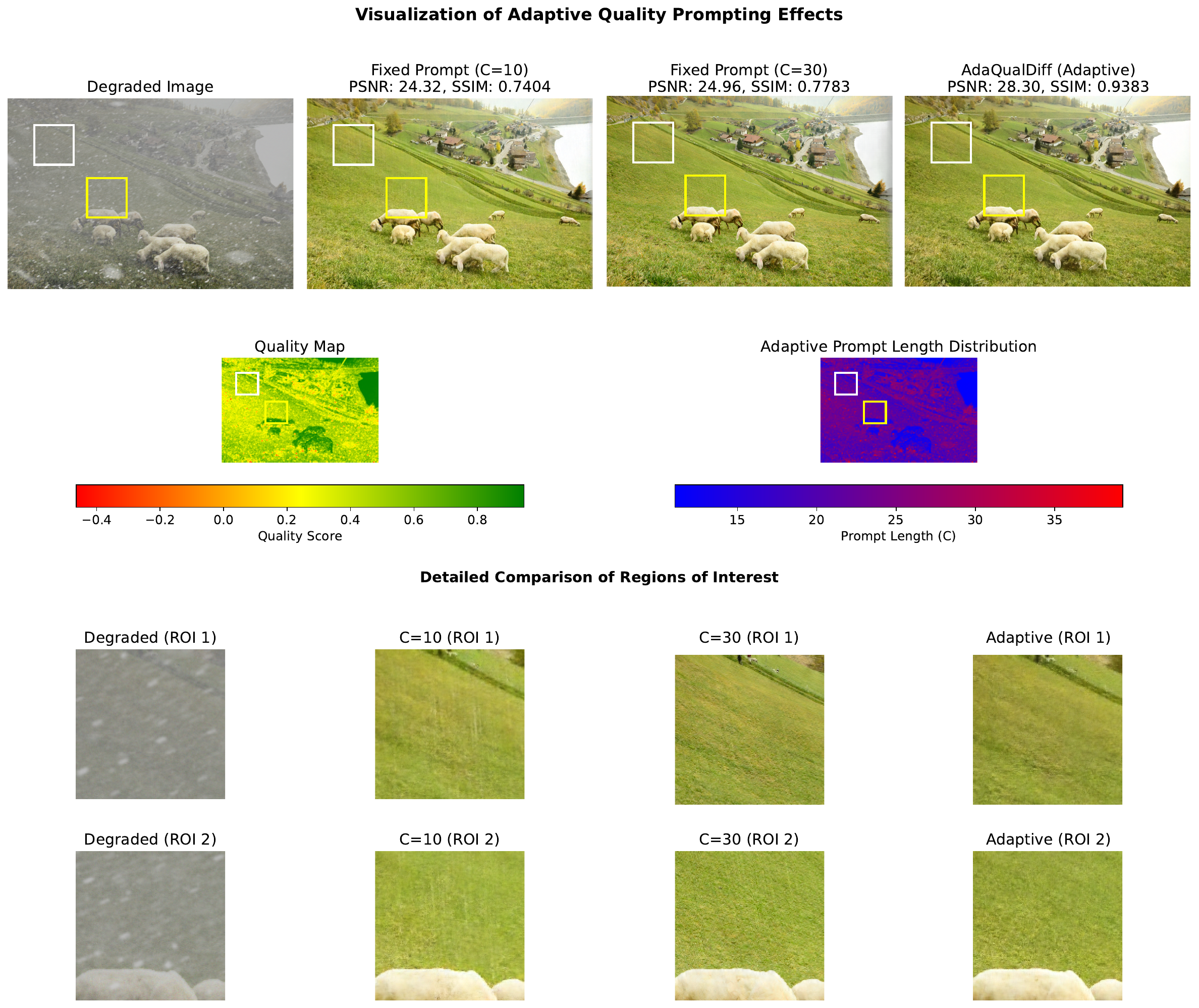}\vspace{-4mm}
	\caption{Adaptive Quality Prompting analysis. (a) Input. (b-d) Comparison of fixed-length (C=10, C=30) vs. our adaptive restoration. (e) Quality heatmap and (f) resulting adaptive prompt allocation.}\vspace{-5mm}
	\label{fig:adaptive_prompting}
\end{figure}

\noindent\textbf{Effectiveness of Adaptive Quality Prompting.} The quantitative results in Table~\ref{tab:ablation_adaptive} demonstrate that fixed-length prompting leads to significant performance degradation. Figure~\ref{fig:adaptive_prompting} provides an intuitive explanation for this phenomenon. Observing the severely degraded regions highlighted by white boxes, short prompts (C=10) fail to provide sufficient guidance for complex degradation scenarios, resulting in inadequate detail recovery. While longer prompts (C=30) perform better in such regions, they introduce unnecessary modifications and potential artifacts in high-quality areas marked by yellow boxes. Our AdaQualDiff method achieves adaptive prompt length allocation, as shown in Figure~\ref{fig:adaptive_prompting}(f), through quality distribution analysis illustrated in Figure~\ref{fig:adaptive_prompting}(e). This mechanism ensures longer prompt sequences are applied to low-quality regions for thorough restoration, while using minimal necessary prompt lengths in well-preserved areas, thereby achieving optimal balance across the entire image. The comparison between the quality heatmap and prompt length distribution clearly reveals how this adaptive strategy effectively allocates computational resources, enabling the restoration process to address severely degraded regions while preserving original details in high-quality areas, thus achieving the optimal performance demonstrated in Table~\ref{tab:ablation_adaptive}.

\begin{table}[!b]\footnotesize 
	\centering
	\caption{Ablation study on the effectiveness of adaptivequality prompting on the CDD11 dataset. Our approach dynamically assigns prompt lengths based on local image quality, leading to optimal performance compared to fixed-length alternatives.}
	\label{tab:ablation_adaptive}\vspace{-4mm}
	\begin{tabular}{lcc}
		\toprule
		\rowcolor{gray!20} 
		\textbf{Method} & \textbf{PSNR (dB)$\uparrow$} & \textbf{SSIM$\uparrow$} \\
		\midrule
		Fixed Prompt (C=10) & 29.33 & 0.8845 \\
		Fixed Prompt (C=20) & 29.42 & 0.8876 \\
		Fixed Prompt (C=30) & 29.87 & 0.8934 \\
		\midrule
		AdaQualDiff (Adaptive) & \textbf{30.11} & \textbf{0.9001} \\
		\bottomrule
	\end{tabular}\vspace{-6mm}
\end{table}

\noindent\textbf{Effectiveness of Quality Assessment Module.}
The effectiveness of AdaQual-Diff relies heavily on the quality assessment that guides its adaptive prompting mechanism. To investigate this dependency, we conduct experiments comparing different quality assessment approaches for generating the spatial quality maps $Q(\mathbf{y})$ that drive our system. We evaluate the following variants: (i) No quality assessment, where we use fixed-length prompts across the entire image; (ii) Compare2Score \cite{zhu2024adaptive}, which uses a pairwise comparison-based quality estimator; (iii) Q-Align \cite{wu2023qalign}, which aligns visual and language representations for quality assessment; and (iv) our full model using DeQAScore.
As shown in Table~\ref{tab:ablation_quality_assessment}, the DeQAScore-based model achieves superior performance, underscoring the importance of accurate perceptual quality assessment for effective restoration guidance. 
\begin{table}[htbp]
	\centering
	\caption{Ablation study on the effectiveness of different quality assessment methods on the CDD11 dataset. The results demonstrate the superior performance of DeQAScore in guiding adaptive prompt complexity.}
	\label{tab:ablation_quality_assessment}
	\begin{tabular}{lcc}
		\toprule
		\rowcolor{gray!20} 
		\textbf{Quality Assessment Method} & \textbf{PSNR (dB)$\uparrow$} & \textbf{SSIM$\uparrow$} \\
		\midrule
		w/o. Quality Assessment & 29.09 & 0.8798 \\
		Compare2Score& 29.58 & 0.8892 \\
		Q-Align & 29.87 & 0.8944 \\
		\midrule
		AdaQual-Diff (DeQAScore) & \textbf{30.11} & \textbf{0.9001} \\
		\bottomrule
	\end{tabular}
\end{table}

\noindent\textbf{Effectiveness of Loss Functions.}
able~\ref{tab:ablation_loss} demonstrates how different loss terms affect restoration performance on the CDD11 dataset. The base configuration employing only L1 loss achieves reasonable results but lacks perceptual quality. Incorporating prompt loss (\ch) yields a +0.43dB PSNR improvement by enforcing content consistency. Further enhancement comes from our quality-weighted prompt mechanism (\ch), which adaptively allocates loss weights based on degradation severity. The full loss configuration, combining all components with our adaptive strategy, delivers optimal performance—validating the effectiveness of our quality-aware approach in complex degradation scenarios.

\begin{table}[!t]
	\centering
	\caption{Ablation study on the effectiveness of different loss function components on the CDD11 dataset.}
	\label{tab:ablation_loss}\vspace{-3mm}
	\resizebox{\linewidth}{!}{
	\begin{tabular}{lccc|cc}
		\toprule
		\rowcolor{gray!20} 
		\textbf{Loss Configuration} & \textbf{L1} & \textbf{Prompt} & \textbf{Quality-weighted} & \textbf{PSNR (dB)$\uparrow$} & \textbf{SSIM$\uparrow$} \\
		\midrule
		Base Loss (L1 only) & \ch & & & 29.35 & 0.8862 \\
		Base + Prompt Loss & \ch & \ch & & 29.78 & 0.8925 \\
		Base + Quality-weighted Prompt Loss & \ch & \ch & \ch & 29.92 & 0.8958 \\
		\midrule
		Full Loss (Ours) & \ch & \ch & \ch & \textbf{30.11} & \textbf{0.9001} \\
		\bottomrule
	\end{tabular}}\vspace{-3mm}
\end{table}

\vspace{-2mm}
\section{Comparison of Parameters and Complexity}
AdaQual-Diff demonstrates significant computational efficiency. Its parameter count of 61.12M represents a reduction of over 50\% compared to diffusion models like IR-SDE and Refusion, and nearly 50\% relative to WeatherDiffusion (Table \ref{tab:com_para}). Crucially, the model requires only 2 sampling steps, substantially lowering the overall computational burden typical of diffusion processes. This efficiency is enabled by our adaptive quality prompting mechanism, which dynamically allocates computational resources based on regional degradation severity, thus minimizing redundancy while preserving high restoration fidelity.
Regarding inference speed (Table \ref{tab:inference_time}), AdaQual-Diff achieves a latency of 17.012ms per image on a single RTX3090 GPU. This performance is highly competitive among composite restoration methods, only marginally exceeding the regression-based OneRestore (16.081ms). This translates to a throughput of approximately 58.8 images per second at batch size 1, underscoring the model's efficiency, particularly notable for a diffusion-based architecture. Such throughput highlights its suitability for practical deployment scenarios where processing speed is critical.

\begin{table}[!t]
	\centering
	\caption{Comparison of parameters and GFLOPs (256$\times$256 resolution) for diffusion process and regressive models.}\vspace{-3mm}
	\begin{tabular}{lcc}
		\hline
		\rowcolor{gray!20} 
		\textbf{Method} & \textbf{\#Params} & \textbf{\#GFLOPs} \\
		\hline
		\multicolumn{3}{l}{\textbf{Image Restoration (Regression)}} \\
		\hline
		Restormer & 25.3M & 140.92G \\
		\hline
		\multicolumn{3}{l}{\textbf{Image Restoration (Diff)}} \\
		\hline
		IR-SDE  & 135.3M & 119.1G$\times$100 steps \\
		Refusion & 131.4M & 63.4G$\times$50 steps \\
		\hline
		\multicolumn{3}{l}{\textbf{Adverse Weather Restoration (Diff)}} \\
		\hline
		WeatherDiffusion & 113.68M & 248.4G$\times$25 steps \\
		T$^3$-DiffWeather & 69.38M & 59.82G$\times$2 steps \\
		\hline
		\multicolumn{3}{l}{\textbf{Composition Degradation Image Restoration (Diff)}} \\
		\hline
		AdaQual-Diff (Ours) & 61.12M & 94.31G$\times$2 steps \\
		\hline
	\end{tabular}\label{tab:com_para}
\end{table}

\begin{table}[htbp]
	\centering
	\caption{Inference Time Comparison of Image Restoration Methods}
	\label{tab:inference_time}
	\renewcommand{\arraystretch}{1}
	\setlength{\tabcolsep}{12pt}
	\begin{tabular}{@{}l|c|c@{}}
		\toprule
		\rowcolor{gray!15} 
		\textbf{Method} & \textbf{Category} & \textbf{Inference Time (ms)} \\ 
		\midrule
		Fourmer & Uniform & 31.590 \\
		MPRNet & Uniform & 75.314 \\
		DGUNet & Uniform & 39.724 \\
		MIRNetv2 & Uniform & 41.391 \\
		\hline
		WGWSNet & All-in-One & 69.746 \\
		PromptIR & All-in-One & 94.330 \\
		\hline
		OneRestore & Composite & 16.081 \\
		OneRestore* & Composite & 16.081 \\
		AdaQual-Diff (Ours) & Composite & 17.012 \\
		\bottomrule
	\end{tabular}
\end{table}
\vspace{-4mm}
\section{Evaluating Real-World Performance.}
\begin{figure}[!t]
	\centering
	\includegraphics[width=\linewidth]{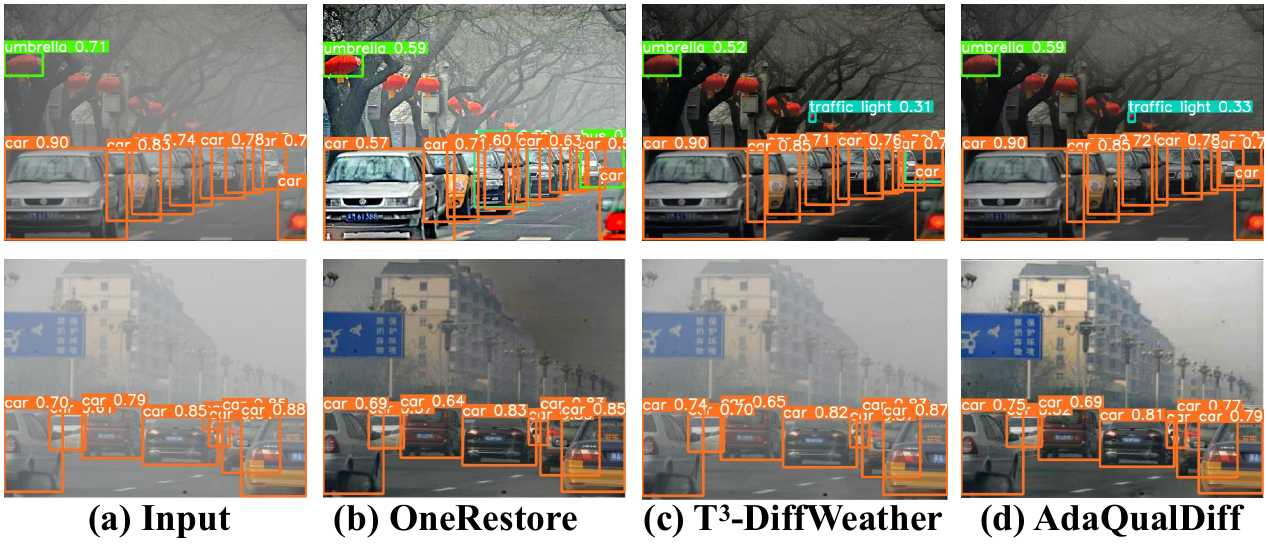}
	\caption{Object detection performance in adverse weather conditions using YOLOv5.}
	\label{fig:real-world}
\end{figure}

\vspace{-1mm}
As shown in Figure \ref{fig:real-world}, we show subjective but quantifiable results for a concrete demonstration, in which the labels and corresponding confidence scores validate the effectiveness of our approach. The figure illustrates how our method successfully handles different degrees of degradation across various environmental conditions, providing both visual quality improvements and maintaining semantic information critical for downstream tasks. Quantitative analysis reveals that object detection confidence scores improve on haze-affected regions, with particular gains observed for small objects and partially occluded instances that benefit most from our quality-adaptive restoration.

\section{Conclusion}
We developed AdaQual-Diff, an image restoration framework optimizing the quality-efficiency trade-off via adaptive quality prompting and a quality-weighted loss within an efficient diffusion model (61.12M parameters, 2 sampling steps). AdaQual-Diff demonstrates robust handling of spatially varying artifacts and yields considerable quantitative improvements compared to baseline methods, particularly for severely degraded regions, confirming the practical utility of targeted quality adaptation strategies.

\bibliographystyle{ACM-Reference-Format}
\bibliography{main}
\newpage
\appendix
\section{Overview}
This supplementary material provides extended details on our AdaQual-Diff framework, comprehensive experimental validation, and additional analysis not included in the main paper due to space constraints. The supplementary content is organized as follows:
\begin{itemize}
	\item Section \ref{sec:supp_data}: Detailed information about dataset configurations.
	\item Section \ref{sec:supp_arch}: Architecture of the AdaQual-Diff framework.
	\item Section \ref{sec:supp_qual}: Additional Qualitative Results.
	\item Section \ref{sec:ablation}: Ablation Study
	\item Section \ref{sec:limitation}: Limitation discussion.		
	\item Section \ref{sec:fea}: Future works.
\end{itemize}


\section{Dataset Configurations}
\label{sec:supp_data}
\noindent \textbf{Composite Degradation Dataset.}
To evaluate the model's capability in handling multiple simultaneous degradations, we adopt the publicly available CDD-11 dataset~\cite{guo2024onerestore}. 
This dataset contains 13,013 degraded/clear image pairs for training and 2,200 pairs reserved for testing. It encompasses 11 distinct degradation categories: low-light, haze, rain, snow, and their various combinations (low+haze, low+rain, low+snow, haze+rain, haze+snow, low+haze+rain, and low+haze+snow).

\noindent \textbf{Adverse Weather Benchmark Datasets.}
Following previous work \cite{10.1007/978-3-031-72673-6_6}, we evaluate our proposed method on several established benchmark datasets targeting specific adverse weather conditions.

\noindent\textbf{Snow100K \cite{liu2018desnownet}:} Designed for image snow removal, this dataset includes testing subsets Snow100K-S, Snow100K-M, and Snow100K-L (representing light, medium, and heavy synthetic snow) and Snow100K-Real (real-world snowy images). We utilize these subsets for evaluation.

\noindent\textbf{Outdoor-Rain \cite{Li_2019_CVPR}:} This benchmark focuses on the removal of both rain and haze. For quantitative analysis, we employ the Test1 subset, which contains 750 high-definition images characterized by dense rain and realistic haze.

\noindent\textbf{RainDrop \cite{qian2018attengan}:} Featuring images with simulated raindrops on the camera lens, we evaluate our method using the RainDrop-A test subset consisting of 58 images.

\section{AdaQual-Diff Architecture}
\label{sec:supp_arch}
Regarding the diffusion process architecture, AdaQualDiff introduces a distinct and arguably more direct paradigm for adaptive restoration compared to methods like $T^{3}$-DiffWeather. While $T^{3}$-DiffWeather innovatively utilizes a prompt pool to dynamically construct \textit{weather-specific prompts} based on degradation residuals and employs \textit{depth features} for general scene guidance, AdaQualDiff pioneers a strategy fundamentally rooted in \textit{direct perceptual quality assessment}. 
Our core creative contribution, the Adaptive Quality Prompting mechanism, moves beyond selecting from predefined pools based on indirect features. Instead, it establishes a direct mathematical relationship between \textit{local image quality scores} (quantified by DeQAScore) and the required \textit{structural complexity and computational intensity} of the guidance prompt itself (\(C_p \propto f(1-Q)\)). This contrasts significantly with $T^{3}$-DiffWeather's approach, which relies on learned correlations between sub-prompts, degradation types, and scene features. 
AdaQualDiff's ingenuity lies in its ability to generate a \textit{spatially-varying guidance field} where restoration effort is precisely and dynamically allocated based solely on the \textit{measured quality deficit} in each region. This allows for exceptionally fine-grained control over the restoration process, directly addressing the severity of degradation without explicit decomposition into weather types or reliance on auxiliary scene representations like depth maps. This quality-driven adaptivity allows our framework to achieve state-of-the-art results with potentially comparable or fewer parameters. 
\begin{figure}[htbp]
	\centering
	\includegraphics[width=\linewidth]{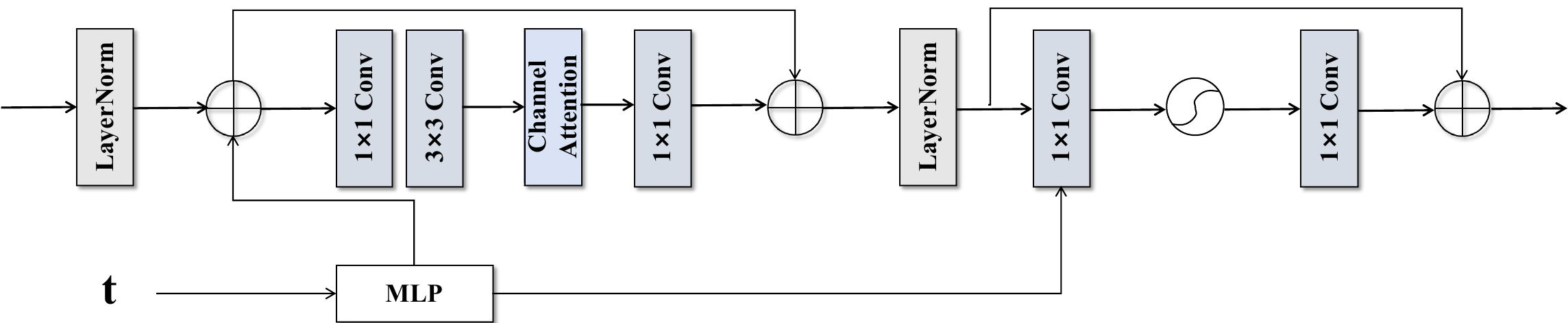}
	\caption{Structure of the NAFNet Block.}\vspace{-4mm}
	\label{fig:structure}
\end{figure}

\section{Additional Qualitative Results}
\label{sec:supp_qual} 

\subsection{More Results on CDD-11 Dataset}
\begin{figure*}[!t]
	\centering
	\includegraphics[width=\textwidth]{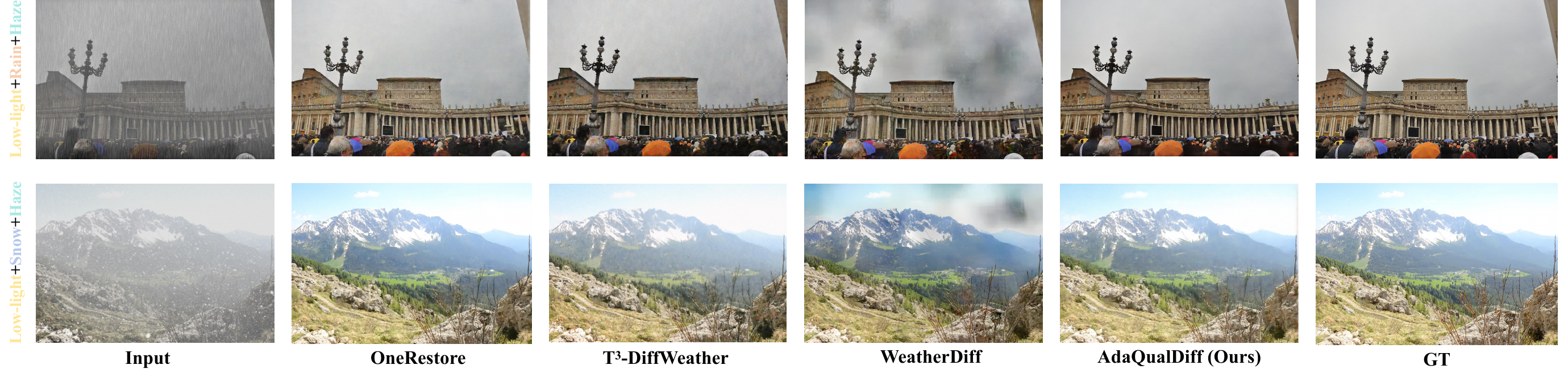}
	\caption{Qualitative comparison on the CDD-11 dataset.}
	\label{fig:cdd_results_1}
\end{figure*}

\begin{figure*}[h!]
	\centering
	\includegraphics[width=\textwidth]{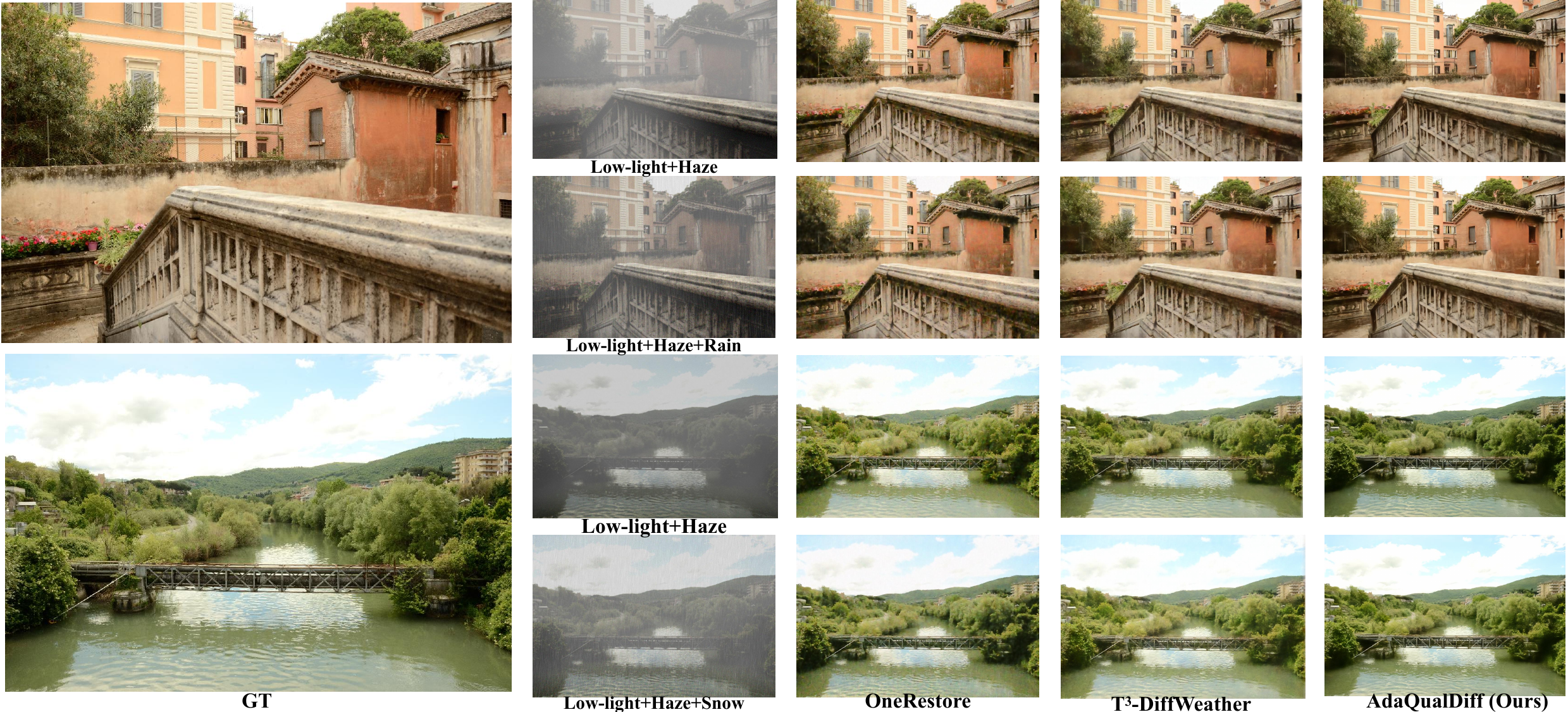}\vspace{-2mm}
	\caption{Qualitative comparison on the CDD-11 dataset.}
	\label{fig:cdd_results_2}
\end{figure*}

\noindent We provide additional visual comparisons on the CDD-11 dataset to demonstrate the effectiveness of AdaQual-Diff in handling complex composite degradations. Figures \ref{fig:cdd_results_1} and \ref{fig:cdd_results_2} (adjust figure references as needed) showcase the results. As observed, AdaQual-Diff achieves superior restoration, producing results closer to the ground truth images compared to existing methods, particularly in recovering fine details and accurate colors under various combined degradations.


\subsection{More Results on Classic Benchmarks}
\noindent We further present extended qualitative results on established benchmark datasets, including Snow100K, Outdoor-Rain, and RainDrop. Figures \ref{fig:adv_results} illustrates these comparisons. 
The figures highlight AdaQual-Diff's ability to generate visually pleasing results that significantly surpass previous methods in terms of overall image detail, texture preservation, and color fidelity across different adverse weather conditions (snow, rain, haze) and raindrop scenarios. 

\begin{figure*}[!t]
	\centering
	\includegraphics[width=\textwidth]{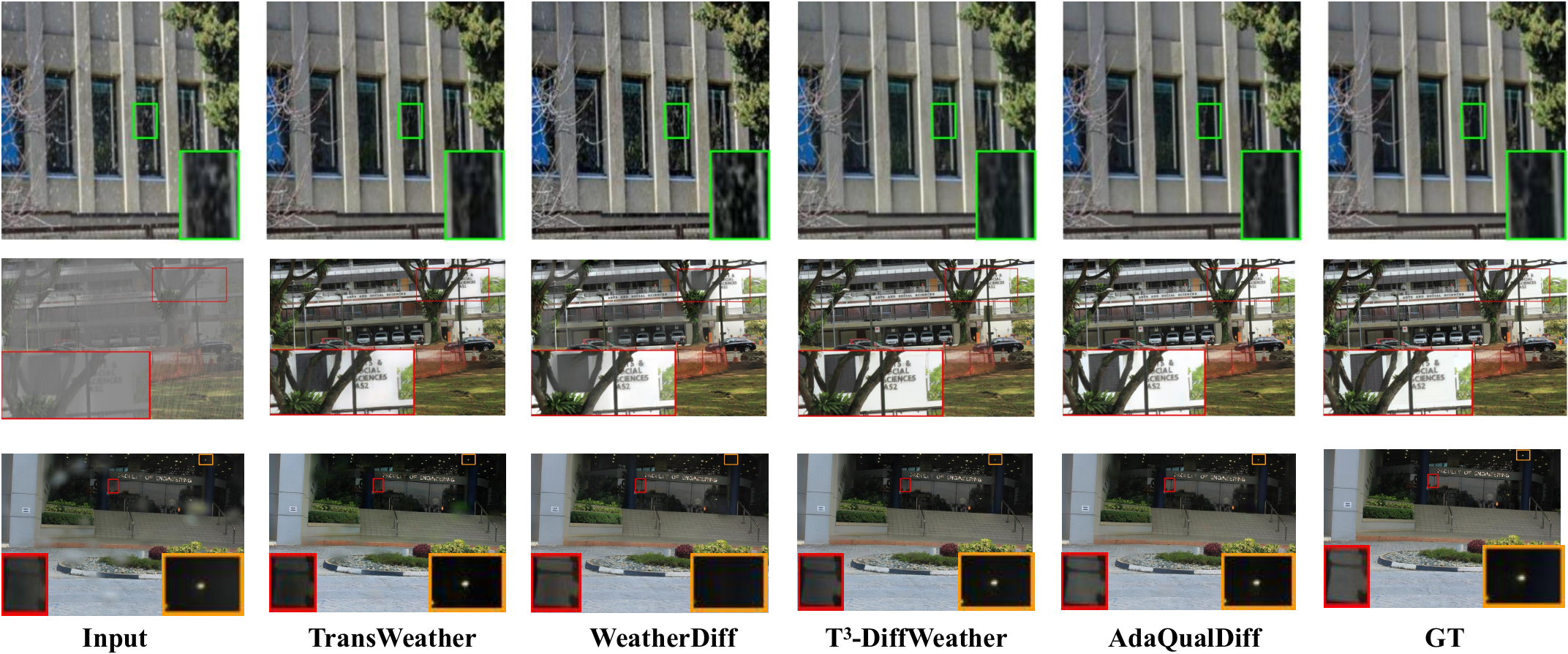}
	\caption{Qualitative comparisons on benchmark datasets Snow100K, Outdoor-Rain, and RainDrop.}
	\label{fig:adv_results}
\end{figure*}
\section{Ablation Study}
\label{sec:ablation}
\noindent In this section, we specifically analyze the sensitivity of the AdaQual-Diff framework to the quality threshold parameter, \(\tau\). This threshold is a crucial element within our Adaptive Quality Prompting mechanism. It determines how the intensity and nature of the restoration guidance (provided via prompts) are modulated based on the input image's perceived quality score \(Q\). This score \(Q\) itself can be informed by advanced assessment methods like DeQA-Score, which notably considers the distribution (mean and variance) of quality scores rather than just a single mean value, highlighting the inherent uncertainty in quality perception.
\begin{table}[h]
	\centering
	\caption{Ablation study examining the effect of the quality threshold \(\tau\) on the restoration performance on the CDD-11 dataset.}
	\label{tab:ablation_threshold}
	\begin{tabular}{c|c c}
		\toprule
		Threshold \(\tau\) & PSNR \(\uparrow\) & SSIM \(\uparrow\) \\
		\midrule
		1.5 (Low Threshold) & 28.91 & 0.878 \\ 
		\textbf{3.0 (Selected)} & \textbf{30.11} & \textbf{0.900} \\ 
		4.5 (High Threshold) & 29.75 & 0.875 \\ 
		\bottomrule
	\end{tabular}
\end{table}

\begin{figure*}[htbp]
	\centering
	\includegraphics[width=0.6\textwidth]{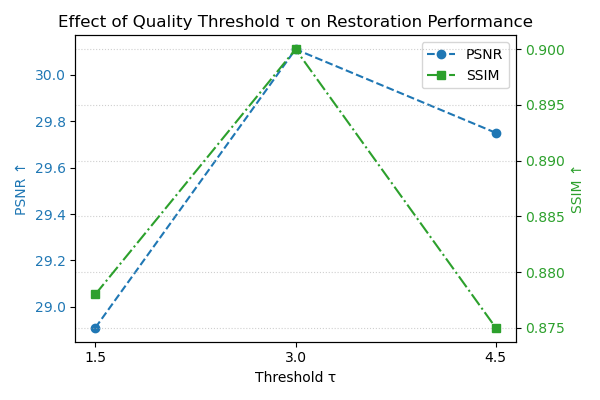}
	\caption{Abl. of the quality threshold \(\tau\).}\vspace{-2mm}
	\label{fig:adv_results}
\end{figure*}
\section{Limitations}
\label{sec:limitation}
Despite the compelling results generated by our proposed pipeline, it also faces some inherent limitations. Firstly, a core characteristic of diffusion models is the significant computational resources required for both training and the iterative inference process. This factor, combined with the model's relatively large parameter count compared to traditional regression methods or other generative frameworks (e.g., VAEs~\cite{jaini2024multiflow,ghosh2024expressive}), potentially hinders deployment in resource-constrained or real-time scenarios. 
\section{Future Work}
\label{sec:fea}
Looking ahead, several avenues warrant further investigation. A key priority is to explore methods for reducing the parameter count of our model, potentially through model compression techniques or architectural refinements, without compromising performance. We also aim to further refine the core mechanisms presented herein, enhancing their effectiveness and robustness. Extending the applicability of our pipeline to encompass a wider range of image restoration tasks within a unified "all-in-one" framework represents a significant future goal. Concurrently, we will continue to scrutinize the pipeline design to eliminate any potential redundancies and optimize overall efficiency.
\end{document}